\documentclass[letterpaper,12pt]{article}
\pdfoutput=1
\usepackage[all]{xy}
\usepackage{amsmath}
\usepackage{amssymb}
\usepackage{amsthm}
\usepackage{amsopn}
\usepackage{mathrsfs}
\usepackage[pdftex]{color,graphicx}
\newtheorem{thm}{Theorem}[section]

\theoremstyle{remark}

\newtheorem{example}[thm]{Example}

\theoremstyle{definition}

\DeclareMathOperator{\Var}{Var}
\DeclareMathOperator{\E}{E}

\newcommand{\ud}{\mathrm{d}}
\numberwithin{equation}{section}
\CompileMatrices
\author{Steven Wegmann \& Larry Gillick \\ Nuance Communications \\ Mobile Research}
\title{Why has (reasonably accurate) Automatic Speech Recognition been so hard to achieve?}
\date{}
\begin{document}
\maketitle 
\section{Introduction}

It has now been over 35 years since hidden Markov Models were first applied to the problem of speech recognition (\cite{baker}, \cite{jelinek}).  Moreover, it has now been over 20 years since the speech recognition community seemed to collectively adopt the HMM paradigm as the most useful general approach to the fundamental problem of modeling speech.   Perhaps a key turning point in this regard was Kai-Fu Lee's thesis work \cite{kflee}, in which he clearly explained how to train an HMM-based system and then successfully applied a series of variations on the HMM theme to the Resource Management task, which was defined by DARPA and where the results were publicly evaluated by NIST.  

This is not to say that there have not been critiques of the HMM as a model of speech, nor that there have not been alternatives proposed and even explored at some length.  One thinks of segmental models of various sorts (\cite{zue}, \cite{ostendorf}), and more recently, of the use of graphical models (\cite{bilmes}). Nonetheless, we think it is fair to say that it is still true in 2010 that the HMM remains the consensus model of choice for speech recognition, and that it lies at the heart of both commercially available products and contemporary research systems.

However, in spite of the great success of the HMM paradigm, all is not well in the Land of Speech Recognition.  Machine error rates on natural speech (e.g. conversational material found in Switchboard or Fisher data) are still very high (around 15\% \cite{chen}), compared to what is achievable by human listeners.\footnote{See the still excellent survey article \cite{lippmann}.  At the time machine error rates were around 40\% on the Switchboard corpus while human performance was around 4\%.}  Moreover, although there has been some progress in the last decade -- mainly through the use of ever larger amounts of acoustic data, in conjunction with a family of training algorithms, typically described as ``discriminative'' -- gains obtained in this manner seem to have slowed down and may well have run their course.

Now it may be argued that the fundamental problem lies with the language model and not with the acoustic model, that human listeners are able to weed out semantically and syntactically implausible transcription alternatives in a way that goes far beyond what our machines can achieve.  While it is certainly true that our language models continue to be downright primitive, we shall argue in this paper that there is an alternative hypothesis that may account for the apparent asymptote in accuracy and that gives some reason for hope that it may be possible to make further significant improvements in accuracy through acoustic modeling.\footnote{The current standard front end, mel filter bank cepstral coefficients (MFCCs), are equally primitive and a likely candidate for improving the accuracy of recognition systems (see, e.g., \cite{morgan}).  However, in this proposal we will only be considering the problem of modeling this admittedly weak feature representation.}  

Let us begin by examining an assumption that lies at the core of hidden Markov Modeling: the statistical independence of frames.  The model makes the very strong assumption that successive frames ``generated'' by a certain state are independent.  Moreover, the model also assumes that the frames generated in one state are statistically independent of those that are generated in a different state, a particularly strong assumption when the states are adjacent in time.  In most cases, acoustic models also assume that individual features within a frame are conditionally independent given the Gaussian that generated them -- here we refer to the standard Gaussian mixture model paradigm in almost universal use in this field -- but, of course, this assumption is different in character.  It is perfectly possible to model the statistical dependence among features in a single frame without violating the standard HMM model.  Our focus in this paper shall fall on the HMM independence assumptions.  It has always, of course, been obvious to everyone that these assumptions are false.  But how false are they?  Is their falsity actually crippling the results we are obtaining? 

Around 10 years ago, we and some of our colleagues at Dragon Systems (\cite{mcallaster}) undertook a series of studies of speech recognition based on the use of simulated data.  Our idea was really derived from a principle that is widely accepted in the field of statistics: if you want to understand the properties of a probabilistic model and of the corresponding estimation procedures, then you should apply them to data created by a known data generation mechanism.  Real speech data, alas, is generated by a mysterious, complex, and unknown process while in those experiments, we generated pseudo utterances from the trained HMMs.  Each such utterance consisted of a frame sequence generated from the relevant parts of the model (a certain number of frames from output distribution 1, followed by a certain number of frames from output distribution 2, etc.), where the series of output distributions is specified by a particular sequence of words, pronunciations for those words, decision trees that pick an output distribution for each state, and so forth.  One conclusion, among others, that we drew from our experiments on pseudo utterances was that if only our real data satisfied the modeling assumptions, our ASR systems would make very few errors.

Back in the late 1970s Bradley Efron, a statistician at Stanford, proposed a general approach called Bootstrapping \cite{efron1}, \cite{efron2} as a method for deriving the properties of statistical algorithms without the necessity of making specific parametric assumptions.  The essence of Efron's idea was that one could generate an unbounded supply of data of the same general type as the original (real) data, by resampling from the original data.  So, for example, if your original data was \(X = ( X_1, X_2, \dots, X_n)\)  and you were interested in the statistic \(S(X)\), one of the basic questions you might have is: what is the variance of \(S\)?  This question could be answered by generating \(M\) replicates of the \(X\) data by resampling (with replacement) from the original \(n\) observations.  For each replicate, say \(Y_1, Y_2, \dots, Y_n\), you can compute \(S\).  Based on the \(M\) values of \(S\), you can then directly estimate \(\Var(S)\) as the sample variance.  In essence Efron proposed that we treat the empirical distribution computed from \(X_1, X_2, \dots, X_n\) as the population distribution.  If \(n\) is large enough, then this empirical distribution should be close to the true population distribution.  Statistical properties computed using the empirical distribution should then be ``close'' to what would have been computed from the unknown population distribution.  

In this paper, we apply Efron's resampling idea to speech data.  Given an utterance U, consisting of a sequence of frames \(X_1, X_2, \dots, X_n\), a speech recognition model M, and a transcription T, consisting of a sequence of words and their corresponding pronunciations, using standard methods we can construct a time alignment between the frames and an HMM state sequence.  For present purposes, let us assume that each frame is assigned to precisely one state, although this is not a requirement -- probabilistic alignments can be handled as well within this framework.  We can repeat this process for a large collection of utterances (namely, all of the test data, for example), and can then collect together all of the frames assigned to each possible HMM state, taking into account whatever state tying has been done, for example by the decision trees.  In the end, then, we have for each distinct output distribution the collection of test frames that are associated with it via a time alignment.  

The next step is to construct a pseudo utterance via resampling.  There are many experiments of this sort that can be done and, indeed, later in this paper we shall illustrate a number of possibilities together with recognition results.    However, first we'll describe a simple example to convey the main idea.  Let us suppose that we single out a particular test utterance, for which we have a time alignment obtained as we have just described.  We therefore know how many frames have been assigned to each state in the sequence by the alignment.  We now proceed to replace each of the original frames by choosing a single frame at random (with replacement) from the inventory of all of the test frames associated with the same state as the original frame.  When we are done with this process, we shall have constructed a resampled utterance of exactly the same length in frames with exactly the same state sequence and alignment.  By construction, however, we shall have ensured that the frames are independent.  This resampled utterance will therefore satisfy the HMM hypotheses.   Moreover, we have done so without imposing any parametric model on the data, unlike in the experiments from a decade ago where we sampled data from a model.  How hard is it to recognize resampled utterances?  How much higher is the error rate on real utterances than on resampled utterances, and by implication, to what extent can statistical dependence account for the corresponding degradation in accuracy?  The experimental results in section~\ref{mainSection} of this paper seek to shed some light on these questions.

The investigations in this paper began as an attempt to understand why discriminative training methods of various sorts have been successful in speech recognition.   Presumably, if our models are correct, then maximum likelihood estimation should be asymptotically efficient  -- in other words, as the sample size increases to infinity, it ought to be impossible to come up with alternative estimates of the unknown parameters that are more accurate on the average.  The natural conclusion, then, has been that discriminative training works because the underlying model is incorrect.  But which aspects of the flawed model is discriminative training correcting for? Moreover, how is discriminative training correcting for those flaws?  After all, discriminative training (whether MMI, or MPE, or MCE, etc.) does nothing to modify the underlying independence assumptions.   It simply leaves the model structure intact, and adjusts the estimates of the parameters.

In one series of experiments we simulated pseudo utterances from the model; we used some of these utterances as training data and the others as test data.  It was not a surprise that when we computed models via maximum likelihood estimation  from the pseudo-training set, and then recognized the pseudo test set, we observed a fairly low error rate.  However, when we then applied discriminative training to the pseudo-training set, and then recognized the pseudo test set with the resulting models, we found to our astonishment that the error rate had decreased.  How could it be that discriminative training had improved the models, when the data had been explicitly generated from a model that satisfied all of the necessary assumptions?

The answer was that we had made a mistake in our experimental setup: as is general practice in speech recognition, we had increased the weight of the language model (LM) score so as to compensate for the presumed dependence in the many individual acoustic scores that were added together.  But in our experiment, there was no dependence among the acoustic scores.  Therefore, the LM score was being over-weighted.  We found that the discriminative training algorithm we were using (maximum mutual information or MMI) was actually able to compensate for the over-weighting of the LM scores by increasing the acoustic scores.  It did so by generally moving the Gaussian means away from the observations that had been averaged to produce the MLE means.  In essence, discriminative training was carrying out what would have been a rather simple global scale adjustment by making many micro adjustments in the means and in the standard deviations.  This inadvertent experiment led us to wonder whether discriminative training methods in speech recognition might generally be partially compensating for dependence among the scores by a subtle collection of parameter modifications that make all scores worse, but by varying amounts.  For example, a vowel with a long duration might lead to a long series of highly correlated scores: the resulting sum of scores might then dominate over the scores from a very short duration consonant.  Perhaps discriminative training could make adjustments to address this kind of phenomenon.  In section~\ref{interestingSection} we shall describe our investigations into this question.

\subsection{Outline}

Here is a brief overview of this paper's structure.  Section~\ref{preliminaries} covers notation, common experimental details, and primers on simulation and resampling using HMMs.  In section ~\ref{interestingSection} we describe experiments that show that MMI can compensate for simple mis-scalings in the acoustic scores.  In section~\ref{simexperiments} we describe our initial simulation and resampling experiments.  In these experiments we create pseudo utterances that obey the HMM generation model but violate the diagonal normal output assumption in a controlled way.  In section~\ref{mainSection} we describe our experiments that study the nature of the statistical dependence in real data and its effect on recognition performance.  We finish with a discussion in section~\ref{discussion}.  

\subsection{Acknowledgments}

We would like to thank Don McAllaster who not only provided the simulation software that we used in the early phases of this research but also gave us helpful feedback on this paper.  We would also like to thank our colleagues Orith Toledo-Ronen, Jim Wu, and George Zavaliagkos for useful discussions about this research.  Finally, we are grateful to Nuance Communications for supporting this research.

\section{Preliminaries} \label{preliminaries}

\subsection{HMM Notation}

We start with a fairly general definition of an HMM.  Let \(X_1, X_2, \dots, X_n\) be a sequence of observed random \(d\)-dimensional acoustic vectors which are distributed according to \(P^{(n)}_\theta\) with the parameter \(\theta\) living in the parameter space \(\Theta\).\footnote{Typically \(\Theta\) is a convex open subset of \(\mathbb{R}^m\) where \(m\) is very large.}  A probability density function for \(X_1, X_2, \dots, X_n\), \(f_\theta(x_1, x_2, \dots, x_n)\), is a \emph{hidden Markov model} if the following three assumptions hold:
\begin{itemize}
\item[(a)] (Hidden chain) We are given, but do not observe, a finite stationary Markov chain \(S_1, S_2, \dots, S_n\) with states \(\{1, 2, \dots, k\}\), stationary initial probability \(\pi_\theta(i)\), \(1 \leq i \leq k\), and \(k \times k\) transition probability matrix \(A_\theta = (a_\theta(i,j))\).
\item[(b)] (Conditional independence) Given \(S_1, S_2, \dots, S_n\) the \(X_i\) are conditionally independent, and given \(S_i\), \(X_i\) is independent of \(S_j\) with \(j \ne i\).  
 \item[(c)] (Stationarity) The conditional distribution of \(X_i\) given \(S_i\) does not depend on \(i\).  We call these \emph{output distributions}.
\end{itemize}
The definitions allow us to write
\begin{equation} \label{hmmMarginalization}
f_\theta(x_1, x_2, \dots, x_n) = \sum_{(s_1,s_2, \dots, s_n)} f_\theta(x_1, x_2, \dots, x_n, s_1, s_2, \dots, s_n).
\end{equation}
If we denote the conditional density of \(X_t\) given \(S_t = j\) by \(f_\theta(x_t \mid s_t = j)\), then the model assumptions give the following decomposition for each term in the sum on the right hand side of (\ref{hmmMarginalization}):
\begin{equation} \label {hmmModel}
 f_\theta(x_1, x_2, \dots, x_n, s_1, s_2, \dots, s_n) = \pi(s_1) \prod_{j=1}^{n-1} a_{\theta}(s_j,s_{j+1}) \prod_{t=1}^n f_\theta(x_t \mid s_t). 
\end{equation}

When we use HMMs in speech recognition we add two non-emitting states to the underlying state sequence, one at the beginning and one at the end.  We do this because we use separate HMMs to model distinct sub-phonetic units (triphones in this paper) and we need to string many HMMs together to model a single utterance.  The non-emitting states facilitate this.  We omit the notational and technical details, but the upshot of this change is that the initial probability mass \(\pi\) ends up being placed entirely on the initial non-emitting state, so (\ref {hmmModel}) becomes
\begin{equation} \label {hmmModel2}
 f_\theta(x_1, x_2, \dots, x_n, s_1, s_2, \dots, s_n) = \prod_{j=1}^{n-1} a_{\theta}(s_j,s_{j+1}) \prod_{t=1}^n f_\theta(x_t \mid s_t). 
\end{equation}

In this paper we shall use particularly simple output distributions, and we shall refer to this as the \emph{diagonal normal output assumption}:
\begin{itemize}
\item[(d)] Each output distribution is multivariate normal with diagonal covariance. 
\end{itemize}

We shall use the abbreviations \(X = (X_1, X_2, \dots, X_n)\), \(x = (x_1, x_2, \dots, x_n)\), \(S = (S_1, S_2, \dots, S_n)\), and \(s = (s_1, s_2, \dots, s_n)\).  We shall consistently use \(t\) for the frame index, \(i\) for the dimensional index, and use them in the order \(x_{t,i}\) to indicate the \(i^\textrm{th}\) component of the \(d\)-dimensional \(t^\textrm{th}\) frame vector \(x_t\).

Let \(W\) be the random transcription taking values in \(\mathcal{W}\), \(p(w)\) is the language model, and \(f_\theta(x \mid w)\) is the acoustic model (the HMM).  Note that we write \(f_\theta(x \mid w)\) instead of just \(f_\theta(x)\) because, as a practical matter, a transcription \(w\) restricts the possible state sequences \(s\) in the sum in (\ref{hmmMarginalization}) to those compatible with \(w\) via the pronunciations.  For any \(\textrm{w} \in \mathcal{W}\), let \(S_{\textrm{w}}\) denote the set of hidden state sequences, \(s\), that are compatible with transcription \(\textrm{w}\) and have the same number of frames as \(x\), namely \(n\).\footnote{For example, in this paper we will be using HMMs to model triphones.  So to construct the set \(S_{\textrm{w}}\) we first take all the phone-level pronunciations consistent with \(\textrm{w}\), then expand them to produce corresponding triphone level pronunciations, and finally enumerate the all of the state sequences with length \(n\) consistent with all of the possible triphone pronunciations.}  Then we define \(f_\theta(x \mid \textrm{w})\) by
\begin{equation*}
f_\theta(x \mid \textrm{w}) = \sum_{s \in S_{\textrm{w}}} f_\theta(x, s).
\end{equation*}
When performing recognition, instead of using the joint distribution \(f_\theta(x, w)\), we use \(f_\theta(x \mid w) p(w)^\kappa\).  The scale \(\kappa\), which is known as the \emph{language model scale}, is used in all practical recognition systems to balance the relative weights of the probabilities obtained from the language model and the acoustic model.  The recognition task is given by
\begin{equation*}
\arg \max_{w \in \mathcal{W}} f_\theta(x \mid w) p(w)^\kappa.
\end{equation*}

\subsection{Experimental preliminaries} \label {experimentsetup}

In this section we give the details that all of our experiments share.  We chose to work on a standard Wall Street Journal (WSJ) task from the early 1990's because, by modern standards, it is small enough so that experimental turnaround is fast even with MMI, but it is large enough so that the results are believable.  This task is also self-contained, with nearly all of the materials necessary for training and testing available through the LDC, the exception being a dictionary for training and testing pronunciations.  We use pronunciations created at VoiceSignal Technologies (VST) using 39 non-silence phones. We use version 3.4 of the HTK toolkit to train and test our models.

We use the WSJ SI-284 set for acoustic model training.  This training set consists of material from 84 WSJ0 training speakers and from 200 WSJ1 training speakers.  It amounts to approximately 37000 training sentences and 66 hours of non-silence data.  Each session was recorded using two microphones; we use the primary channel recorded using a Sennheiser microphone.

The VST front-end that we use produces a 39 dimensional feature vector every 10 ms: 13 Mel-cepstral coefficients, including c0, plus their first and second differences.   The cepstral coefficients are mean normalized.  The data is down-sampled from 16 kHz to 8 kHz before the cepstral coefficients are computed.

We use very small, simple acoustic models.  Not only does this lower the computational load for the experiments involving MMI, but it also helps by making certain effects easier to see: real data should be more surprising to these smaller, simpler models, than they would be to larger more sophisticated models.  The acoustic models use word-internal triphones.  Except for silence, each triphone is modeled using a three state HMM without skipping.  For silence we follow the standard HTK practice that uses two models for silence: a three state tee-model and a single state short pause model; the short pause model is tied to middle state of the longer model; both models allow skipping.\footnote{See \cite{young} for details.} The resulting triphone states were then clustered using decision trees to 1500 tied states.  The output distribution for each tied state is a single, multivariate normal distribution with a diagonal covariance.

We report word error rate (WER) on two test sets.  Using the nomenclature of the time, these test sets use the 5k closed vocabulary and non-verbalized punctuation.  The first test set is the November 1992 ARPA evaluation 5k test set.  It has 330 sentences collected from 8 speakers.  The second test set is referred to as si\_dt\_05.odd in \cite{woodland1}.  It is a subset of the the WSJ1 5k development test set defined by first deleting sentences with out of vocabulary (OOV) words (relative to the 5k closed vocabulary) and then selecting every other sentence.  This results in 248 sentences collected from 10 speakers.  Together, these two test sets amount to about an hour of non-silence data.  We test using the standard 5k bigram language model created at Lincoln Labs for the 1992 ARPA evaluation.  The combined WER rate on these test sets using the models described above is 18\%.

When we refer to MMI training, we mean lattice-based extended Baum-Welch as described in \cite{povey2} or \cite{woodland2}.  We use HTK 3.4 to perform extended Baum-Welch with standard settings, e.g, \(E = 1\).  We use a VST tool and a relatively weak bigram language model trained from the acoustic training sentences (we kept bigrams that had 8 or more examples) to generate word lattices on the training set.  We use HTK tools to create phone-marked numerator and denominator lattices, the latter starting from the word lattices described above.  The minimum WER of 12\% occurs on our combined test set after 10 passes of extended Baum-Welch.

\subsection{Simulating a pseudo utterance from an HMM} \label{simprimer}
In this section we give a simulation primer.  The classic reference for this material is \cite{abramowitz}, chapter 26.
 
Here is an overall description of how we generate a parallel version of a corpus by simulating from an HMM.  We leave the actual simulation details in (brief) abeyance.  The inputs are the model, a dictionary, the transcript, and the real utterance.  We use the model, the transcript, the dictionary, and forced alignment to pick the pronunciations (we discard the time information) and inter-word silence (sp or sil).  We use an alignment to guarantee that the real and pseudo utterance share the underlying triphone sequence, but a perfectly reasonable alternative is to make random selections among the pronunciations and silence types.\footnote{If we are creating pseudo utterances from scratch, i.e., if we do not have extant utterances, then by necessity we randomly select the pronunciations.} This results in a triphone sequence for the utterance, e.g., sil a+b a-b+c b-c sil.  Next we generate the underlying state sequence for the utterance.  For each triphone in the list we generate a state sequence by simulating from the transition models. This produces a list of state id's : 1 1 2 2 3 4 \ldots, and determines the number of frames.  Finally we generate the actual frames: we walk down the list of state id's, simulating one frame from the corresponding output distribution.

\begin{figure}
\centering
\includegraphics[width=.9\textwidth]{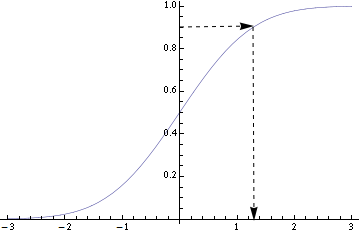}
\caption{Simulating from \(N(0,1)\).\label{normalsim}}
\end{figure}

Now we turn to the details of simulation.  How do we simulate data from a continuous distribution?  The following result provides the key:
\begin{thm} \label {contsimthm}
Let \(F\) be a continuous, invertible, cumulative distribution function and let the continuous random variable \(U\) have uniform distribution on \([0,1]\).  If we define the continuous random variable \(X\) by \(X = F^{-1}(U)\), then \(X\) has distribution \(F\) 
\end{thm}
To use Theorem \ref {contsimthm} to generate data having distribution \(F\), first we use a random number generator to choose \(u \in [0,1]\), then we find the unique \(x\) satisfying \(F(x) = u\).  Figure \ref{normalsim} illustrates this procedure on \(N(0,1)\), with \(u=0.9\) and the resulting value \(x = 1.25\).  The next two examples show how the simulation procedure for \(N(0,1)\) is used to simulate multivariate normal distributions: first with diagonal covariance and second with full covariance.  

\begin{example}\label{normalsimex} {Simulating from a \(d\)-dimensional normal distribution with diagonal covariance.} Let \(\Phi\) be the cumulative distribution for \(N(0,1)\), and let \(\mu\) and \(\sigma^2\) be the \(d\)-dimensional mean and diagonal variance vectors.\footnote{When we write \(\sigma^2\) we mean the vector with components \(\sigma^2_i\).}  We use a random number generator \(d\) times to create the \(d\)-dimensional vector \(u\) with each component \(u_i \in [0,1]\) and use this to create a \(d\)-dimensional vector \(y\) by the rule \(y_i = \Phi^{-1}(u_i)\) for \(i \leq i \leq d\).  Then the vectors \(x = \mu + y^t \sigma\) will have distribution \(N(\mu, \sigma^2)\).  
\end{example}

\begin{example} {Simulating from a \(d\)-dimensional normal distribution with full covariance.} Let \(\mu\) \(d\)-dimensional mean and \(\Sigma\) be a positive definite \(d \times d\) covariance matrix.  Since \(\Sigma\) is positive definite there exists a non-singular \(d \times d\) matrix \(M\) with \(\Sigma = M M^t\).\footnote{The matrix \(M\) is not unique, but the Cholesky decomposition is particularly useful for this application.} If we follow the procedure in Example \ref{normalsimex} to create the vectors \(y\), then the vectors \(x = \mu + M y\) will have distribution \(N(\mu, \Sigma)\).  
\end{example}

How do we simulate data from a discrete distribution?  We use a discrete analog of Theorem \ref{contsimthm}, that we describe in the following example:
\begin{example}{Simulating from a discrete probability distribution.}\label{discreteSim}  Let the discrete probability distribution \(P\) have probabilities \(\{p_i\}_{i=1}^n\) with \(n>1\) and that satisfy \(0 < p_i < 1\) for each \(i\) and \(\sum_{i=1}^n p_i = 1\).  Define intervals \(\{A_j\}_{j=1}^n\) by
\begin{equation*}
A_j = \left\{ \begin{array}{ll}
[0,p_1) & \textrm{if \(j = 1\)}\\
\textrm{\([\sum_{i=1}^{j-1} p_i, \sum_{i=1}^{j} p_i)\)} & \textrm{if \(1 < j < n\)}\\
\textrm{\([\sum_{i=1}^{n-1} p_i, 1]\)} & \textrm{if \(j = n\)}.
\end{array} \right.
\end{equation*}
Then the \(\{A_j\}_{j=1}^n\) form a partition of \([0,1]\), which means that \(\cup_{j=1}^n A_j = [0,1]\) and \(\cap_{j=1}^n A_j = \emptyset\).  We use this partition to define a function \(h: [0,1] \to \{1,2,\dots,n\}\) in the following way: given \(u \in [0,1]\) there is a unique \(j\) with \(u \in A_j\) and we set \(h(u) = j\). Finally, if the continuous random variable \(U\) has uniform distribution on \([0,1]\), then we define a discrete random variable \(X\) on a set of values \(\{x_i\}_{i=1}^n\) by \(X = h(U)\).  By construction \(X\) has distribution \(P\), i.e. \(P(X=x_i) = p_i\).
\end{example}

\begin{figure}
\centering
\includegraphics[width=.9\textwidth]{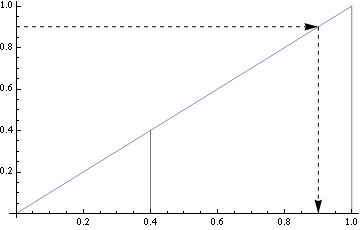}
\caption{Simulating a Bernoulli random variable with \(p=0.4\).\label{geometricsim}}
\end{figure}

\begin{example}\label{bernoulliSim} To help clarify Example~\ref{discreteSim}, we work through the details in the special case \(n=2\).  We can simplify the notation by setting \(p = p_1\) from which it follows that \(1-p = p_2\).  The two intervals \(A_1\) and \(A_2\) are given by \(A_1 = [0, p)\) and \(A_2 = [p, 1]\).  The function \(h\) is given in terms of \(u \in [0,1]\) by
\begin{equation*}
h(u) = \left\{ \begin{array}{ll}
1 & \textrm{if \( u \in [0,p)\)}\\
2 & \textrm{if \( u \in [p,1] \)}\\
\end{array} \right.
\end{equation*}
Since \(U\) has uniform distribution over \([0,1]\), it follows that the probability distribution of the random variable \(X = h(U)\) is given by
\begin{equation*}
P(X=1) = P(u \in [0,p)) = p
\end{equation*}
and
\begin{equation*}
P(X=2) = P(u \in [p,1]) = 1-p.
\end{equation*}
To simulate the random variable \(X\) we first use a random number generator to choose \(u \in [0,1]\).  We next set the value of \(X\) to be 1 if \(1 \leq u < p\) and 2 otherwise.  Figure \ref{geometricsim} displays this simulation procedure in a manner that is analogous to Figure~\ref{normalsim} when \(p = 0.4\) and \(u = 0.9\).  Finally, we note that in this case, \(n=2\), we can think of \(X\) as a Bernoulli random variable that takes value 1 with success probability \(p\) and 2 with failure probability \(1-p\).  
\end{example}

\begin{example}\label{geometricSim} {Simulating from a geometric distribution.}  The geometric distribution with parameter \(p\), \(0 < p < 1\), is a discrete probability distribution with probabilities \(\{p_i\}_{i=1}^\infty\) given by\footnote{It is a standard exercise to show that \(\sum_{i=1}^\infty p_i = 1\).}
\begin{equation*}
p_i = (1-p)^{i-1} p.
\end{equation*}
Since the geometric distribution is discrete -- albeit infinite -- it is straightforward to directly adapt the machinery described in Example~\ref{discreteSim} to simulate data from it.  However, we will describe an alternate procedure that is based on the intuition that a geometric random variable is the waiting time for a success in a series of independent Bernoulli trials.  To do this, we first introduce a Bernoulli random variable, \(X\), that takes value 1 with success probability \(p\) and value 0 with failure probability \(1-p\).  We then conduct a series of independent Bernoulli trials, \(X_1, X_2, \dots\), until we obtain a success.  If we let \(Y\) be the random variable that gives the number of trials needed for the first success, then \(Y\) has a geometric distribution with parameter \(p\), since
\begin{equation*}
P(Y = n) = P((X_1, X_2, \dots, X_n) = (0, 0, \dots, 1)) = (1-p)^{n-1} p = p_n.
\end{equation*}
Thus to simulate one example from a geometric distribution, it suffices to simulate a series of Bernoulli trials until we obtain a success.  Example~\ref{bernoulliSim} shows that this is equivalent to repeatedly running a random number generator, which takes values in \([0,1]\), to create a sequence \(u_1, u_2, \dots\)  The first \(n\) satisfying \(u_n \in [0, p)\) gives the required value of \(Y\).   
\end{example}

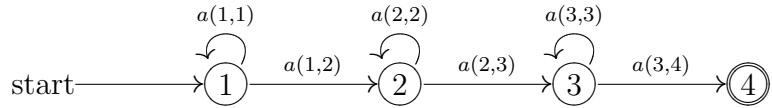
\begin{figure}
\centering
\begin{displaymath}
\entrymodifiers={++[o][F-]}
\SelectTips{cm}{}
\xymatrix @C+=4pc {
*\txt{start} \ar[r]
& 1 \ar[r]^{a (1,2)} \ar@(ur,ul)[]_{a (1,1)} 
& 2 \ar[r]^{a (2,3)} \ar@(ur,ul)[]_{a (2,2)}
& 3 \ar[r]^{a (3,4)} \ar@(ur,ul)[]_{a (3,3)}
&*++[o][F=]{4} }
\end{displaymath}
\caption{A linear three state HMM without skipping.\label{hmmNoSkip}}
\end{figure}

Figure~\ref{hmmNoSkip} displays the linear transition structure that we typically use in HMMs for speech recognition.  However, the transition structure in the HMM displayed in Figure~\ref{hmmNoSkip} also forces the model to spend at least one frame in each state, i.e., no state skipping is allowed.  To simulate a state sequence from the model displayed in Figure~\ref{hmmNoSkip} we apply the method described in Example~\ref{geometricSim} to each of the three states.  For example, for the first state we need to simulate the waiting time for the states sequence to move on to the second state.  In this case the Bernoulli random variable, \(X\), describes if we stay in state 1 with failure probability \(a(1,1)\) or we move on to state 2 with success probability \(a(1,2)\).

We end this primer with a brief remark.  The frames that we use in these experiments use a feature set that contains the first and second differences of the cepstral features.  When we create a pseudo utterance by simulating from the HMM, the resulting frame sequence is somewhat peculiar in the sense that the difference features are not consistent with the underlying sequence of static cepstral features.  This is because the HMM knows nothing about this structure, in fact, this structure violates the model assumptions.  In section~\ref{ceponlysection} we examine this more carefully.

\subsection{Resampling from a corpus} \label {resampleprimer}

In this section we briefly describe how resampling differs from simulation in the context of an HMM.  The reader should consult \cite{efron2} for more details.  We shall use resampling to generate frames from the output distributions.  At the heart of the resampling method is a collection of labeled urns, one for each state in the HMM, that contain frames.  When we need a frame from a state, we choose a single frame (with replacement) from the corresponding urn.

How do we fill the urns with frames?  We start with a collection of utterances that have state level alignments obtained from the HMM either using (a) forced alignment or (b) forward-backward.  We walk through all of the frames in the collection, putting each frame in its corresponding urn with count one in case (a) and a fractional count (the posterior probability that the frame was generated by the state) in case (b).

How do we label the urns?  The are two labeling schemes that we use in this paper, which we call \emph{speaker dependent} and \emph{speaker independent} resampling.  We use speaker dependent resampling if we want the pseudo utterances to preserve the speaker labels in the real utterances.  To do this we create speaker specific urns, and put frames in the urn with the correct speaker and state label.  To do speaker independent resampling we ignore the original speaker labels when filling the urns.  A pseudo utterance created using speaker independent resampling may contain frames from possibly all of the speakers in the collection of utterances, while a pseudo utterance created using speaker dependent resampling contains frames from only one speaker.

How do we create a pseudo utterance?  We use the procedure described in section~\ref{simprimer}, except when it is time to get a frame from the state, instead of simulating a frame from the output distribution, we randomly select a frame from the appropriately labeled urn.  The remaining technicality that we need to address is how do we draw a frame from an urn containing fractional counts?  We create a cumulative distribution using the fractional counts, simulate from this distribution, and then select the corresponding frame.  

\section{Experiments involving MMI and score scales} \label{interestingSection}

In this section we start by giving a more detailed description of the anomalous results involving the LM scale that was outlined in the introduction.  We then show that the MMI machinery can compensate for simple artificial scalings that we introduce into pseudo data. 

\subsection{An anomaly}

We created pseudo test and training data by simulating from the model.  When we ran recognition on the pseudo test data with the LM scale set to 16, the WER was 2.0\%, which seemed reasonably low at the time.  We then generated lattices on the pseudo training data, again with the LM scale set to 16, and ran 100 passes of extended Baum-Welch.  To our surprise the WER steadily decreased during the passes of extended Baum-Welch: starting from 2.0\% at the mle, it decreased to 1.3\% after 10 passes, and ended at 0.7\% after 100 passes.  As noted in the introduction, we knew that something was awry because the test and training data were simulated from the model.  How could any other choice of parameters lead to better classification performance?\footnote{One possibility that we needed to eliminate was that there was a bug in our code!  As a sanity check, we trained maximum likelihood models from the pseudo training data and verified that these models were indistinguishable from the models that we simulated the pseudo training data from.}

Of course the problem was that we were using LM scale 16.  This choice is what we have routinely used for WSJ experiments for many years, but it is the wrong choice when the data matches the model!  When performing recognition, instead of using the joint distribution \(f_\theta(x, w)\), we use \(f_\theta(x \mid w) p(w)^\kappa\), where \(\kappa\) is the LM scale.  The LM scale is related to how much we distrust our acoustic model and when the data is simulated from the model it should be 1 because the acoustic model is correct and we should be using the joint distribution.

When we reran recognition on the pseudo test data with the correct scale, \(\kappa = 1\), we found that the WER using the mle was much lower, namely 0.2\%.  Moreover when we reran recognition on the pseudo test data, again with \(\kappa = 1\), but on the models produced using extended Baum-Welch on lattices generated using \(\kappa = 16\) on the pseudo training data, the WER error gets steadily worse, ending at 8.9\% after 100 passes.

We then regenerated the lattices on the pseudo training data this time with \(\kappa = 1\).  We reran extended Baum-Welch and observe no change in the model parameters from pass to pass, exactly as one would expect. 

\subsection{Is MMI compensating for differences in score scales?}

The results of the previous section are very intriguing: we accidentally introduced an extra scaling (\( 1 / 16\)) that was uniformly applied to the acoustic model scores during extended Baum-Welch and during recognition.  After 100 passes of extended Baum-Welch the WER decreased from 2.0\% to 0.7\% compensating for much of the degradation that this scaling introduced.

It should not be a surprise to most readers that the degree to which our models fit the data varies widely among the states.  For example, it seems implausible that data will fit the diagonal normal assumption to same degree from state to state, or that frames associated with silence are as correlated as those associated with a vowel.  In fact much of the rest of this paper is devoted to quantifying the degree to which the model agrees or disagrees with the data.  One way of thinking about this varying degree of model correctness is that the scores produced by different states are inherently on different scales, whereas our recognizer treats them all as if they are on the same scale. Could it be that MMI is simply tinkering with the model parameters so as to weight the models by different degrees?

\begin{figure}
\centering
\includegraphics[width=1\textwidth]{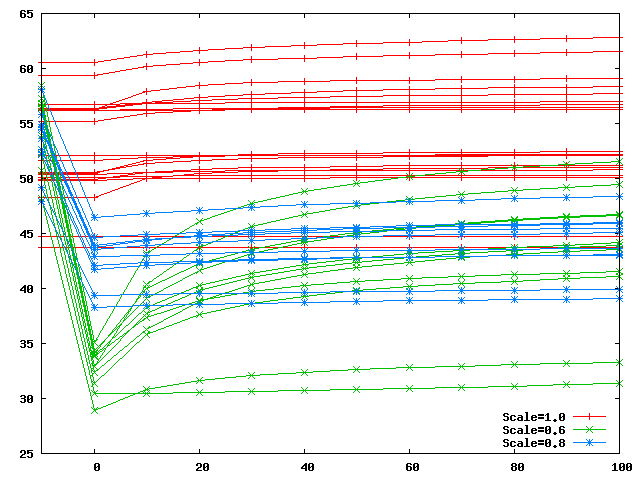}
\caption{Tracks of the average phone scores in the three scale case.\label{scoretracks}}
\end{figure}

Here is a simple experiment to test whether or not MMI is capable of recovering from more complicated score mismatches.  We use pseudo training data simulated from the model and numerator and denominator lattices created from this data with \(\kappa = 1\).  We run extended Baum-Welch and recognition on the lattices, the latter by finding the best path through the merged numerator and denominator lattices.  However, we shall multiply the acoustic score of each phone arc by a phone dependent scale.  We shall try three types of scales:
\begin{itemize}
\item [(a)] No mismatch, with all scales = 1.
\item [(b)] Uniform mismatch, all scales \(= 1/ 16\).
\item [(c)] Variable mismatch with three phone dependent scales.  Silence and vowels have scale = 1.  Half of the remaining phones have scale = 0.8 while the other half have scale 0.6.
\end{itemize}
We have seen cases (a) and (b) before, but not in the context of lattice re-scoring.  The results are displayed in Table \ref {scaleTable}.  MMI is remarkably good at compensating for these very simple scalings.  Figure \ref {scoretracks}
 shows how the average scores for each phone change: the scores along the y-axis are the original unscaled scores, the scores at x = 0 are after the initial scaling, and the subsequent labels along the x-axis give the extended Baum-Welch pass.  Even though it may not be obvious from inspection, all of the scores get worse as extended Baum-Welch progresses, even the scores from phones that were unscaled.  However, as it is clear from inspection, the scores from the phones with scale = 0.6 move the most.  It is worth noting that even though extended Baum-Welch does not restore the scores from the phones with scale 0.6 and 0.8 to anywhere near their original values, the WER nearly returns to it original value. 

\begin{table}
\centering
\begin{tabular}{|r|c|c|c|}
\hline
pass & no scale & scale \(= 1 / 16\)  & three scales\\
\hline
\hline
0 & 0.84 & 4.58 & 32.40 \\
10 & 0.84 & 2.60 & 3.81 \\
20 & 0.84 & 2.04 & 1.96 \\
30 & 0.84 & 1.77 & 1.41 \\
40 & 0.84 & 1.61 & 1.25 \\
50 & 0.84 & 1.47 & 1.16 \\
60 & 0.84 & 1.38 & 1.07 \\
70 & 0.84 & 1.31 & 1.02 \\
80 & 0.84 & 1.25 & 0.99 \\
90 & 0.84 & 1.20 & 0.96 \\
100 & 0.84 & 1.16 & 0.93\\
\hline
\hline
\end{tabular}
\caption{Word error rates as MMI progresses with three types of scaling.\label{scaleTable}}
\end{table}  

\section{Initial experiments with simulated and resampled data} \label {mainfirstexperiments}

In this section we shall begin by describing our initial experiments using simulated test and training data.  At the time these results were both surprising and discouraging.  They made sense only after we turned to experiments using resampled data.

These experiments are all aimed at trying to understand to what degree real data deviates from the diagonal normal output assumption and what effect this has on recognition performance.  In this section we shall demonstrate that real data do significantly depart from the form of our output distributions but that this departure does not appear to be a major cause of errors.

\subsection{Simulation experiments} \label {simexperiments}

Our idea at the time was to use simulation to create training and test data that deviated from the model in controlled ways.  We would then run the MMI machinery to see if and how it compensated for the known mismatch.  There are two obvious model assumptions that we start with, both involving the diagonal normal output assumption, namely, our use of diagonal instead of full covariance and our use of the normal distribution.  Our use of unimodal models should make it easier to see any effects.

As a baseline we created pseudo test data by simulating from this model and recognizing: the WER is 0.2\% (16 errors).  One conclusion that we can make directly from this result is that there is virtually no overlap in the 1500 output distributions that we are using, since if there were any appreciable overlap we would have seen a higher error rate.  We find this to be very surprising!

The first experiment tests whether or not MMI is compensating for the diagonal covariance assumption, since real data clearly violate this assumption and other researchers have speculated that one of the main effects of MMI is to compensate for this specific example of data/model mismatch.  To test this, first we will create pseudo test data that from an HMM that has full covariance, normal output distributions.  To estimate these models, we simply re-estimate the covariances for the each of the states in our baseline HMMs by accumulating the statistics for full covariances instead of the usual diagonal covariances.  We accomplish this using Baum-Welch training starting from the baseline HMMs and only updating the covariances -- e.g. we do not modify the means.  We then created pseudo test data by simulating from the new HMM with full covariance, normal output distributions, then recognized this test data with the original diagonal covariance models: the resulting WER is 0.3\% (25 errors).  This a disappointingly small change in the error rate.

\begin{figure}
\centering
\includegraphics[width=.9\textwidth]{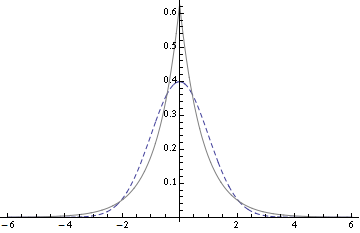}
\caption{Comparison of \(N(0,1)\) (dashed) and \(L(0, \sqrt{2 / \pi})\) densities.\label {laplacevsnormal}}
\end{figure}

Real data have distributions with heavier tails than the normal distribution.  In the second experiment, we shall use the Laplace or double exponential distribution to try explore how important this issue is.  We denote the Laplace distribution with location parameter \(a\) and scale parameter \(b\) by \(L(a,b)\).  Recall that the maximum likelihood estimates for \(a\) and \(b\) are the median and the mean absolute deviation of the data.  Also recall that if \(X = (X_1, X_2, \dots, X_n) \stackrel{i.i.d}{\sim} N(\mu, \sigma^2)\), then the median of \(X\) is \(\mu\) and the mean absolute deviation of \(X\) is \(\sqrt{2 / \pi} \sigma\).\footnote{The first statement is obvious, while the second follows from \(\int_{-\infty}^{+\infty} |x| \varphi_{(0,\sigma)}(x) \ud x = \sqrt{2 / \pi} \sigma\), where \(\varphi_{(0,\sigma)}\) is the \(N(0, \sigma^2)\) density.}  We use this to convert each of our normal models to Laplace models by replacing each \(N(\mu, \sigma^2)\) by \(L(\mu,\sqrt{2 / \pi} \sigma)\), rather than directly fitting Laplace models to the data.  We then created pseudo test data by simulating from the model with these Laplace output distributions.  Figure \ref {laplacevsnormal} compares a 1-dimensional normal density to the corresponding, converted Laplace density; it also shows how pseudo data simulated from a normal models differs from the pseudo data simulated from a converted Laplace model.   This time the WER is 0.4\% (37 errors), but again it is disappointingly small.

\subsection{The predicted versus observed distribution of acoustic model scores} \label {chisqscores}

We were so surprised by the last results that we began to wonder if some of our assumptions about the data were correct.  In particular how far do the data depart from the unimodal normal output distributions?  To study this question, instead of studying the frames themselves, we use the corresponding acoustic model scores.  First we introduce a bit of notation.  We use \(j\) to index our 1500 states, whose diagonal normal output distributions have 39-dimensional means \(\mu_j\) and variances \(\sigma_j^2\).  We shall use \(i\) to index the 39 dimensions, and denote the \(i^{\textrm{th}}\) component of the 39-dimensional vector \(\mu_j\) by \(\mu_{j,i}\).  The acoustic score of state \(j\) of a frame \(x_t\) is, modulo constants, given by
\begin{equation} \label{scoreDef}
V_j(x_t) = - \log f_\theta(x_t \mid s_t = j) = \frac{1}{2} \sum_{i=1}^{39} \frac{(x_{t,i}-\mu_{j,i})^2}{\sigma_{j,i}^2} + \frac{1}{2} \sum_{i=1}^{39} \log \sigma_{j,i}^2.
\end{equation}

If our model were correct, then frames, \(X\), emitted by state \(j\) would satisfy \(X \sim N(\mu_j, \sigma^2_j)\).  If, as usual, we let \(\chi^2_{39}\) denote the chi-squared distribution with 39 degrees of freedom, then
\begin{equation*}
2 V_j - \frac{1}{2} \sum_{i=1}^{39} \log \sigma_{j,i}^2 \sim \chi^2_{39}.
\end{equation*}
We can now use standard properties of \(\chi^2_{39}\) to compute the expected value and variance of \(V_j\) under the assumption that \(X \sim N(\mu_j, \sigma^2_j)\):
\begin{equation*}
\E(V_j) = \frac{39}{2} + \frac{1}{2} \sum_{i=1}^{39} \log \sigma_{j,i}^2
\end{equation*}
and
\begin{equation} \label{alignvar}
\Var(V_j) = \frac{39}{2}.
\end{equation}

We would like to use (\ref{alignvar}) in a hypothesis test with the null hypothesis being that the HMM with diagonal normal output distributions generated the data. However, to use (\ref{alignvar}) we would need to know which state each frame is assigned to.  However in our model the underlying state sequence, \(S\), is hidden.  So we either have to pick this most likely underlying state sequence, i.e. force align, or assign each frame fractional counts across all the states using Forward-Backward.  We choose to do the latter.  During a pass of Baum-Welch training we keep track of all of the acoustic scores emitted by the states.  For each training frame, \(x_t\), and state \(j\) we store \(V_j(x)\) and \(p_\theta(s_t = j \mid x)\) which is the fraction of the frame \(x_t\) occupied by state \(j\) -- this is often referred to as the occupancy of state \(j\) at time \(t\).  Let \(F_{\theta}\) denote the distribution of frames across the states that is defined by the collection of probabilities \(\{p_\theta(s_t = j \mid x)\}_{t,j}\).  We use the fractional counts to compute \(\E_{F_{\theta}}(V_j)\) and \(\Var_{F_{\theta}}(V_j)\) using
\begin{equation} \label {evdef}
\E_{F_{\theta}}(V_j) = \frac{\sum_t p_\theta(s_t = j \mid x) V_j(x_t) }{ \sum_t p_\theta(s_t = j \mid x)}
\end{equation}
and
\begin{equation*}
\Var_{F_{\theta}}(V_j) =  \frac{\sum_t p_\theta(s_t = j \mid x) (V_j(x_t) - \E_{F_{\theta}}(V_j))^2 }{ \sum_t p_\theta(s_t = j \mid x)}.
\end{equation*}
If the data were generated by the model and if we also assume that we have run Baum-Welch enough times to guarantee that our parameter updates have converged to \(\hat{\theta}\), then the cloud of points \(\{x_t\}_t\) with fractional assignments to state \(j\) will be distributed \(N(\hat{\mu}_j, \hat{\sigma}^2_j)\).  Our previous analysis applies again to show that under these assumptions:
\begin{equation} \label {evcompute}
\E_{F_{\hat{\theta}}}(V_j) = \frac{39}{2} + \frac{1}{2} \sum_{i=1}^{39} \log \hat{\sigma}_{j,i}^2
\end{equation}
and
\begin{equation} \label {devcompute}
\Var_{F_{\hat{\theta}}}(V_j) = \frac{39}{2}.
\end{equation}

Recall that at convergence the update equations for the means and variances for each state \(j\) are given by:
\begin{equation} \label {meanformula}
\hat{\mu}_j = \frac{\sum_t p_{\hat{\theta}}(s_t = j \mid x) x_t }{ \sum_t p_{\hat{\theta}}(s_t = j \mid x)}
\end{equation}
and
\begin{equation} \label {devformula}
\hat{\sigma}^2_j = \frac{\sum_t p_{\hat{\theta}}(s_t = j \mid x) (x_t -  \hat{\mu_j})^2}{ \sum_t p_{\hat{\theta}}(s_t = j \mid x)}.
\end{equation}
It follows from (\ref {evdef}), (\ref {meanformula}), and  (\ref {devformula}) that, for each state \(j\), (\ref {evcompute}) holds whether or not the training data fit the model.  So (\ref {evcompute}) is not useful for testing our hypothesis.\footnote{However, (\ref{evcompute}) is useful for testing the null hypothesis Baum-Welch has converged. In our experiments the data do support this hypothesis.}  On the other hand, the update equations (\ref {meanformula}) and  (\ref {devformula}) do not imply that (\ref {devcompute}) holds, so we may use this to test the hypothesis that the HMM with diagonal normal output distributions generated the data.

\begin{figure}
\centering
\includegraphics[width=.9\textwidth]{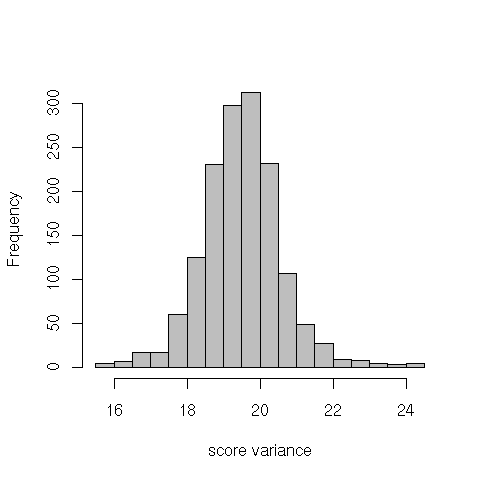}
\caption{Histogram of score variances on simulated data.\label {scoresynthvarhist}}
\end{figure}

First we use pseudo training data simulated from the model to verify that if the data are normal, then (\ref {evcompute}) holds.  Figure \ref {scoresynthvarhist} displays a histogram of the collection \(\{\Var_{F_{\hat{\theta}}}(V_j)\}_{j=1}^{1500}\) computed from this pseudo training data.  The distribution of score variances is approximately normal with median = 19.5 in good agreement with what one would expect.

\begin{figure}
\centering
\includegraphics[width=.9\textwidth]{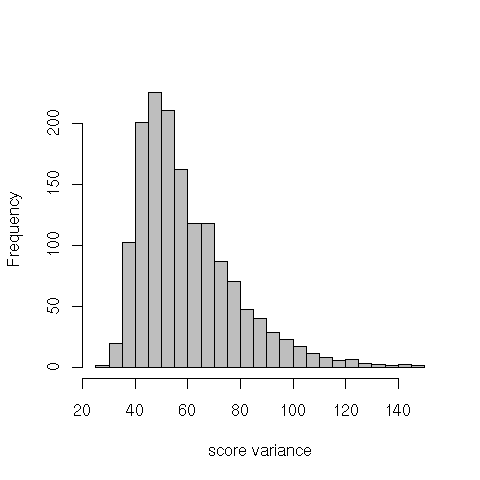}
\caption{Histogram of score variances on real data.\label {scorevarhist}}
\end{figure}

\begin{figure}
\centering
\includegraphics[width=.9\textwidth]{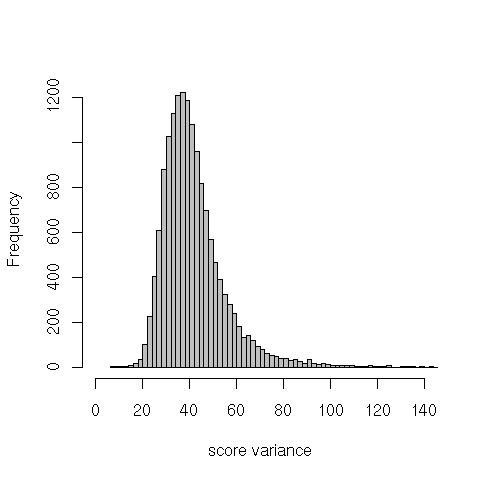}
\caption{Histogram of score variances on real data with 10 component mixture models.\label {scorevarhist10comp}}
\end{figure}

Next we examine the distribution of score variances computed from real training data, which is shown in Figure \ref {scorevarhist}.  This distribution is very different from what we would expect if the diagonal normal output assumption were true.  In particular, the median of the observed score variances is 54.9 which is more than double than the median under the null hypotheses, 19.5, and in fact the minimum observed score variance is 29.3.  Clearly the data strongly departs from the model assumption.  Could it be that our unimodal models make this departure artificially worse?  To answer that, we used mixture models with 10 diagonal normal components for the output distributions, and computed the per component variance in the scores.  Figure \ref {scorevarhist10comp} displays a histogram of the collection of per component variances \(\{\Var_{F_{\hat{\theta}}}(V_j)\}_{j=1}^{15000}\) computed from real training data.  Again the distribution of score variances departs from what we would expect if the model were correct.

\begin{figure}
\centering
\includegraphics[width=.9\textwidth]{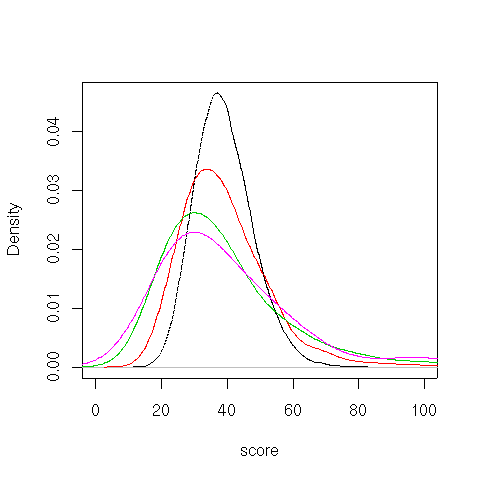}
\caption{Comparison of the distributions of the scores emitted by three triphone states and \(\chi^2_{39}\). Grey is \(\chi^2_{39}\), red is state 2 from ah-p+iy, green is state 3 from ao-r+ey, \& violet is state 2 from r-ax+r. \label {chisquared}}
\end{figure}

We take a closer look at the acoustic scores emitted by three triphones in Figure \ref {chisquared}.
\begin{itemize}
\item [(a)] ah-p+iy state 2, score variance = 42.
\item [(b)] ao-r+ey state 3, score variance = 93
\item [(c)] r-ax+r state 2, score variance = 91
\end{itemize}
To facilitate comparison with \(\chi^2_{39}\) we plot
\begin{equation*}
2 V_j - \frac{1}{2} \sum_{i=1}^{39} \log \hat{\sigma}_{j,i}^2 \sim \chi^2_{39}.
\end{equation*}
We see that not only do these score distributions differ from \(\chi^2_{39}\), but they differ among themselves.

We have shown that the real data do not fit the normal output distribution assumption.

\subsection{Resampling experiments}

In section~\ref{simexperiments} we created three sets of pseudo test data by simulating from three different HMMs.  The first HMM is our unaltered baseline HMM, while the second and third HMMs differ from the baseline only in their output distributions: the baseline HMM has diagonal multivariate normal output distributions, the second HMM has full covariance output distributions, and the third HMM has diagonal Laplacian output distributions.  We observed that the WERs obtained by using the baseline HMM to recognize the three sets of pseudo test data are remarkably similar (0.2\%, 0.3\%, and 0.4\% respectively) and dramatically lower than the corresponding WER on the real test data (18\%).  We can consider these results in the context of a more general problem: we create pseudo test data by simulating from the baseline HMM's Markov model to create the underlying state sequence, but when we create the frames instead of simulating from the baseline HMM's output distributions we simulate from alternate output distributions.  When we recognize this pseudo test data with the baseline HMM, how different can these alternate output distributions be from the baseline HMM's diagonal, multivariate normal distributions for the resulting WER to remain low?  The results of ~section~\ref{simexperiments} suggest that we will observe low error rates largely independent of the form of the alternative output distributions.  If we could verify this claim somehow, then it would imply that WERs are high on real data not because of its departure from the form of the baseline HMM's output distributions but because of its departure from the baseline HMM's generation model, i.e. the conditional independence of frames.  Unfortunately, it seems rather difficult to formulate this claim precisely, let alone verify it.  To get around this difficulty, in this section we shall use resampling to verify a related distribution-free claim that is more clearly related to the problem at hand: if we create pseudo test utterances using resampling, that is we follow the method outlined in section~\ref{resampleprimer} by
\begin{itemize}
\item[(a)] creating the state sequence by simulating from the baseline HMM's Markov model and
\item[(b)] creating the frames by randomly selecting frames (with replacement) from the appropriate state's urn,
\end{itemize}
then recognition WERs on this pseudo test data using the baseline HMM is very low.  Thus we will have demonstrated that if real data satisfied the HMM's independence assumptions, then WERs using the baseline HMM would be dramatically lower.  This is in spite of the fact that, as we saw in section~\ref{chisqscores}, real data does significantly depart from the diagonal normal output distributions in the baseline HMM.

We performed two speaker independent resampling experiments.  In both experiments we fill our state urns with frames that have fractional counts obtained from forward-backward.  In the first experiment we create pseudo test utterances from urns filled from the training data while in the second experiment we fill the urns from the test data.  In both cases the WER obtained by recognizing the resampled data using the baseline HMM is 0.5\%, with 44 and 50 errors respectively.  These error rates are similar to the WER (0.2\%) obtained by using pseudo test data simulated from the baseline HMM, i.e. pseudo test data that satisfies all of the baseline HMM's assumptions and all of these WERs are dramatically lower than the WER using the real test data (18\%).

\begin{figure}
\centering
\includegraphics[width=.9\textwidth]{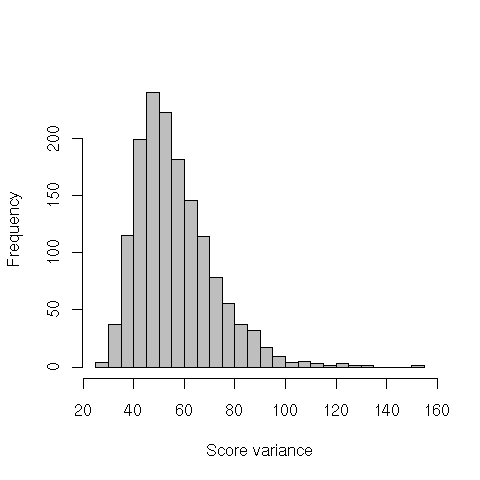}
\caption{Histogram of score variances on resampled training data.\label {resamplescorevarhist}}
\end{figure}

As a simple sanity check on our resampling code, we decided to verify that the resampled data really do violate the diagonal normal output assumption to the same degree that real data do. When we draw a frame in the resampling procedure we sample from a cumulative distribution obtained from the HMM, so it is possible that this distribution puts very low probability on outliers, effectively excluding them.  To check that this does not happen, we use the hypothesis test from section~\ref{chisqscores}.  We create pseudo training data using speaker independent resampling and use it to compute the score variances \(\{\Var_{F_{\hat{\theta}}}(V_j)\}_{j=1}^{1500}\).  Figure \ref {resamplescorevarhist} shows a histogram of these score variances which is very similar to what we obtained on real data (Figure \ref {scorevarhist}).  The resampled data really does violate the diagonal normal output assumption to the same degree as real data. 

To finish the section, let's review our results.  We created two pseudo test sets using speaker independent resampling -- one pseudo test set used the training data to fill each state's urn, and the other used the original, real test data to fill each state's urn.  In both cases the resampled test data satisfies all of the model assumptions except for the diagonal normal output assumption and in both cases the recognition WER obtained using the baseline HMM is very low (0.5\%).  In section~\ref {chisqscores} we verified that the data are not normal, yet we see from these results that this non-normality does not lead to a large change in WER.\footnote{Granted there is a 68\% relative improvement in the WER as you move from pseudo test data created by resampling from the test data to pseudo test data created by simulating from the model, but the error rates involved are tiny.}  The results in this section are consistent with the results from section~\ref {simexperiments}.  We have shown that pseudo test data that satisfy the underlying Markov assumption exhibit low WERs whether or not they satisfy the diagonal normal output assumption.  Put another way, if real data satisfied the Markov assumption, then violation of the diagonal normal output assumption could not be responsible for the degradation in WER from test data that fits our model (0.2\%) to real test data (18\%).  Finally, when we compare WER on the original real test data to WER on the resampled pseudo data, the WER drops from 18\% to 0.5\%.  It follows that the real test data does not satisfy the Markov assumptions, and that this violation is a big source of recognition errors.

\section{Controlling for statistical dependence} \label{mainSection}

All of the simulation and resampling experiments in section~\ref {mainfirstexperiments} created pseudo utterances that satisfy the underlying Markov assumption in our model, or put another way, these pseudo utterances respect the generation model of the HMM.  In this section we describe a series of novel experiments that create pseudo utterances that violate this generation model.  Not only do we finally see significant word error rates on these pseudo utterances, but we demonstrate that statistical dependence is a major cause of recognition errors on real data.  We also investigate the source of the dependency by analyzing the correlation structure in the scores emitted by the models.  Finally we present two variants of the main experiment that, in the first case, explores how much of the dependency we observe is due to the correlation that is built into our feature set, and in the second case, presents preliminary results using speaker dependent resampling. 

\subsection{The main experiment} \label {mainsubsection}
First we force align the real test data with the true transcripts to get a state level alignment, i.e., each frame is assigned a unique state id.  Next we use these alignments and speaker independent resampling to create pseudo test utterances. The result of these operations is a pseudo test set parallel to the original, real test set such that each test utterance has a real and resampled version satisfying:
\begin{itemize}
\item [1.] They share a common underlying state sequence.
\item [2.] They have the same number of frames.
\item [3.] Each resampled utterance is a mix of frames taken from possibly all of the test speakers while each real utterance is from one speaker.
\end{itemize}

We partition each utterance into what we call \emph{state regions} based on the underlying state id's, namely each state region is a maximal set of constant state id's.  For example, if an utterance's underlying state sequence is \(1 1 1 1 2 3 3 3 3 3 3 3 \), then we break it into three state regions: \(1 1 1 1\), \(2\), and \(3 3 3 3 3 3 3 \).  In our test set there are 429250 frames distributed over 127729 state regions, so each state region lasts 3.4 frames on average.

We create alternate versions of the test data by performing two types of operations on the frames within each state region:
\begin{itemize}
\item [(a)] Repeat the first frame throughout the rest of the state region.
\item [(b)] Exchange real frames with corresponding resampled frames.
\end{itemize}
Table \ref {operationTable} lists the types of operations that we shall perform along with short-hand, explanatory 'codes'.

\begin{table}
\centering
\begin{tabular}{|l|c|c|r|}
\hline
code & frame source & state region specific modification & WER\\
\hline
\hline
\(r_1 r_2 r_3\) & real & none & 17.7\\
\(r_1 r_1 r_1\) & real & repeat the first frame & 23.8\\
\(s_1 s_2 s_3\) & re-sampled & none & 0.6\\
\(s_1 s_1 s_1\) & re-sampled & repeat the first frame & 4.4\\
\(r_1 s_2 s_3\) & hybrid & replace subsequent frames with re-sampled & 5.7\\
\(r_1 s_1 s_1\) & hybrid & repeat the first re-sampled frame & 9.4\\
\hline
\hline
\end{tabular}
\caption{The state region operations, their codes, and error rates.\label{operationTable}}
\end{table}

Continuing with our simple example, let
\begin{equation*}
(rf_{1}, rf_{2}, rf_{3}, rf_{4}, rf_{5}, rf_{6}, rf_{7}, rf_{8}, rf_{9}, rf_{10}, rf_{11}, rf_{12})
\end{equation*}
be the real frame sequence and
\begin{equation*}
(sf_{1}, sf_{2}, sf_{3}, sf_{4}, sf_{5}, sf_{6}, sf_{7}, sf_{8}, sf_{9}, sf_{10}, sf_{11}, sf_{12})
\end{equation*}
be the resampled frame sequence.  Then the operation with code \(r_1 r_1 r_1\) produces an utterance with frames
\begin{equation*}
(rf_{1}, rf_{1}, rf_{1}, rf_{1}, rf_{5}, rf_{6}, rf_{6}, rf_{6}, rf_{6}, rf_{6}, rf_{6}, rf_{6}),
\end{equation*}
while the operation with code  \(r_1 s_2 s_3\) produces an utterance with frames
\begin{equation*}
(rf_{1}, sf_{2}, sf_{3}, sf_{4}, rf_{5}, rf_{6}, sf_{7}, sf_{8}, sf_{9}, sf_{10}, sf_{11}, sf_{12}).
\end{equation*}

Table \ref {operationTable} also presents recognition results on test sets created using these operations.  When examining these results remember that we created the pseudo utterances by resampling from the test data.  Consequently, the real and pseudo utterances violate the diagonal normal output assumption to the same degree.  In fact the real and pseudo utterances differ only in the order in which the frames are presented.  We shall now use Table \ref {operationTable} to make statements about within state region dependence and between state region dependence.

Note that repeating resampled and real frames within the state regions has markedly different consequences: we see a dramatic drop in the error rate (86 \% relative) when we compare \(s_1 s_1 s_1 \to s_1 s_2 s_3\), and a large but much smaller drop in the error rate (26 \% relative) when we compare \(r_1 r_1 r_1 \to r_1 r_2 r_3\).  In both cases we are multiplying the number of distinct frames by the same factor, namely \(3.4\).\footnote{This is the average number of frames per state region.}  By construction, the resampled frames are independent, while the results show that the additional real frames \(r_1 r_1 r_1 \to r_1 r_2 r_3\) carry much less information than the additional resampled frames \(s_1 s_1 s_1 \to s_1 s_2 s_3\).  It follows that the additional real frames must be statistically dependent and that this dependence is occurring within the state regions.

Next we examine the relative improvements in the following three comparisons:
\begin{itemize}
\item [(a)] 82\% : \(r_1 r_1 r_1 \to s_1 s_1 s_1\)
\item [(b)] 89\% : \(r_1 s_2 s_3 \to s_1 s_2 s_3\)
\item [(c)] 53\% : \(r_1 s_1 s_1 \to s_1 s_1 s_1\)
\end{itemize}
In each case we start with state regions where the leading frame is the only original real frame, and we replace it with a single resampled frame.  By construction the collection of lead resampled frames are independent, so it follows that the collection of leading real frames must be statistically dependent and this dependence is occurring between the state regions.

Case (b) is worth discussing in more detail.  All that we do is substitute the leading real frame from each state region with a resampled frame (this operation changes fewer than one third of the frames), but this drastically reduces the error rate.  In both cases the frames violate the diagonal normal output assumption to the same degree, but the pseudo utterances with code \(s_1, s_2, s_3\) satisfy the HMM generation model, while the pseudo utterances with code \(r_1 s_2 s_3\) have a single frame in each state region that we have shown does not satisfy the HMM generation model.  The removal of the statistical dependence of these frames, which on average are 3.4 frames apart, is the only possible cause for the in the 89\% relative reduction in the WER observed in this case.  This is truly remarkable!

We have studied two possible sources for the dependence: dependence because of speaker effects and dependence across time, e.g. correlations among adjacent frames.  Instead of studying the frames directly, we proceed as in section~\ref{chisqscores} and look at the corresponding acoustic model scores, where the score of a frame is defined by (\ref{scoreDef}).  The scores have the advantage of being one-dimensional and the scores are independent if and only if the underlying frames are independent.  Here is a rough approximation guided by (\ref{scoreDef}): if we compute a correlation coefficient \(\rho\) on each of the features in a collection of frames, then we would expect to compute a correlation of \(\sim \rho^2\) on the corresponding collection of scores.\footnote{We have verified this empirically by creating simulated data with several degrees of correlation imposed uniformly on the features in frames and computing the corresponding correlation coefficients on scores.}  To obtain the acoustic model scores we score each frame against the state id that was obtained by forced alignment and used to produce the state regions.

For each speaker we measure the between state region correlation in the following way: we create an ordered list of the scores emitted be the leading frame of the state regions; we compute the coefficient of correlation of adjacent\footnote{We do not cross utterance boundaries.} scores relative to the mean of the scores in the list.  To try to determine how much of this correlation is due to speaker effects, we re-use the lists of scores, but randomize them and compute the coefficient of correlation relative to the total mean of the scores in all of the lists.  Table \ref {speakercorrwer} presents these correlation coefficients along with the WER for all of the test speakers on the original, real test data.  The scores are quite correlated across the state regions and the correlation coefficient is relatively constant across speakers.  On the other hand, the amount of speaker dependent correlation varies widely across speakers, with only three speakers showing any significant degree of correlation.  The largest correlation coefficient is 0.205 which is weak.  Also, there is no clear connection between this speaker dependent correlation and the corresponding word error rate.

\begin{table}
\centering
\begin{tabular}{|c|r|c|c|r|}
\hline
Speaker & Count & \(\rho\) : Between & \(\rho\) : Speaker & WER\\
\hline
\hline
440 & 8867  & 0.581 & 4.55e-03& 14.3\\
441 & 8824  & 0.643 & 1.45e-01& 22.3\\
442 & 9733  & 0.615 & 2.77e-02& 15.9\\
443 & 9692  & 0.630 & 4.16e-02& 10.6\\
444 & 10094 & 0.572 & 4.95e-03& 19.5\\
445 & 7795  & 0.620 & 2.87e-02& 14.1\\
446 & 9311  & 0.638 & 2.61e-02& 9.6\\
447 & 8987  & 0.569 & 8.57e-03& 18.9\\
4k0 & 5481  & 0.569 & 2.48e-02& 22.5\\
4k1 & 5583  & 0.634 & 5.39e-02& 30.0\\
4k2 & 6181  & 0.637 & 1.78e-02& 12.8\\
4k3 & 5762  & 0.598 & 3.01e-02& 17.3\\
4k4 & 4846  & 0.530 & 2.05e-01& 17.2\\
4k6 & 5547  & 0.641 & 1.20e-02& 21.3\\
4k7 & 5873  & 0.640 & 4.52e-02& 21.3\\
4k8 & 5492  & 0.575 & 4.78e-02& 10.7\\
4k9 & 5193  & 0.614 & 8.82e-03& 23.9\\
4ka & 5046  & 0.614 & 1.81e-01& 28.4\\
\hline
\hline
\end{tabular}
\caption{Correlations.\label{speakercorrwer}}
\end{table}

\begin{figure}
\centering
\includegraphics[width=.9\textwidth]{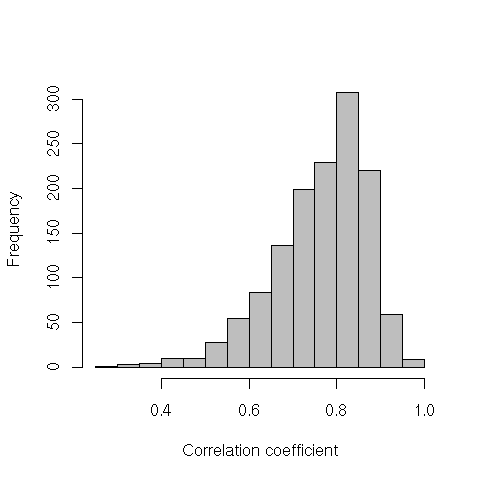}
\caption{Histogram of within state region correlation coefficients on real test data.\label {realwithinhist}}
\end{figure}

\begin{figure}
\centering
\includegraphics[width=.9\textwidth]{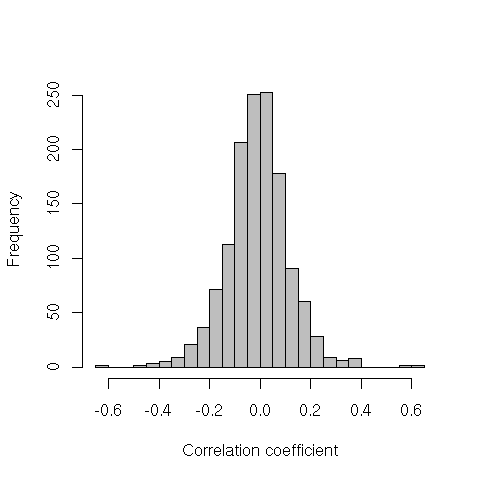}
\caption{Histogram of within state region correlation coefficients on speaker independent resampled test data.\label {siresampledwithinhist}}
\end{figure}

We also computed the within state region correlation coefficients: for each state we computed the coefficient of correlation between adjacent scores emitted in each state region relative to the mean of all of the scores emitted by that state in the test data.  Figures \ref {realwithinhist} and \ref {siresampledwithinhist} display histograms of the correlation coefficients on real and speaker independent resampled test data, where we have excluded states with fewer than 20 examples.  This results in 1351 states for both real and resampled test data.  Adjacent scores within a state region in real test data are highly correlated, while the scores are not correlated in the resampled data.

Before proceeding with two variations on this experiment, we should clarify an important point.  A careful reader might complain about the fact that we are using the HMM to choose the most likely underlying state sequence for the test utterances.  This state sequence is used to (a) create the state regions that we use to show that there is statistical dependence among the frames and (b) to produce the acoustic model score sequence that we show exhibits strong time dependence.  How much do these results, which invalidate the model, depend on the alignment, which depends on the model?  One answer is that we are performing a form of hypothesis testing similar to what we did in section~\ref{chisqscores}.  We examine properties of the data under the null hypothesis that the model generated the data.  Under this null hypothesis, it is perfectly reasonable to ask the model for the state sequence that most likely explains the data.  We reject the null hypothesis, however, because we observe that the data are not conditionally independent, which is a key model assumption.  A second, related answer is that we are using the state sequence to create a somewhat arbitrary partition of the data into what we are calling state regions.  Relative to these state regions, the data exhibit strong statistical dependence that we are able to measure, that we show leads to large WER increases, and violates our model assumptions.  How we created these state regions is irrelevant.

\subsection{Variant 1: cepstral features} \label{ceponlysection}

The 39-dimensional feature set that we use consists of 13 cepstral features plus their first and second differences.  The inclusion of the differences guarantees that adjacent frames are correlated to a certain degree. How much of the time dependence that we observed in section~\ref{mainsubsection} is attributable to this?  To answer this question, we repeat the experiments using only the 13 cepstral features.\footnote{Our front-end uses a 25ms analysis window but generates 100 frames per second, so the resulting overlap means that adjacent cepstral features are correlated.}  To create this data we simply extract the 13-dimensional subspaces from the models and the features.

Table \ref {13vs39errors} compares results on pseudo data created using simulation from the model versus speaker independent resampling using 39 or 13-dimensional features.  Since the error rates are so low, ranging from a minimum of 0.2\% to a maximum of 1.4\%, we show the numbers of errors on these test sets in Table \ref {13vs39errors} rather than WER.  The low number of errors in the simulation results imply that there is surprisingly little overlap in the models in 39 or 13 dimensions.  Likewise, the resampling results show that there is surprisingly little overlap in the test frames in 39 or 13 dimensions.  In both cases the number of errors increases by a factor of \(\sim 3\) as we move from 39 dimensions to 13 dimensions which implies that there must be more separation in the models and the data in the 39-dimensional space.

\begin{table}
\centering
\begin{tabular}{|l|r|r|}
\hline
 & \multicolumn{2}{|c|}{Number of errors}\\ \cline{2-3}
creation method & 39 features & 13 features\\
\hline
\hline
simulation & 16 & 49\\
resampling & 52 & 133\\
\hline
\hline
\end{tabular}
\caption{Comparison of the number of errors on simulated and resampled test data using 39-dimensional features and 13-dimensional features.\label{13vs39errors}}
\end{table}

\begin{figure}
\centering
\includegraphics[width=.9\textwidth]{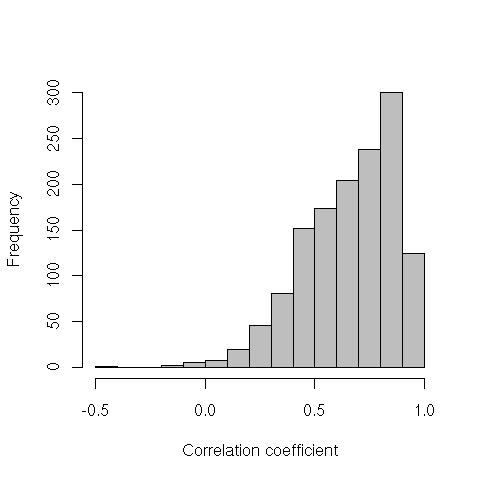}
\caption{Histogram of within state region correlation coefficients on 13-dimensional real test data.\label {realwithinhistcep}}
\end{figure}

Table \ref{compabswercepstral} shows the six operations and their effects on WER on the original 39-dimensional feature set and the 13-dimensional cepstral features.  Table \ref {comprelwercepstral} shows the relative improvements in WER for five comparisons using the original 39-dimensional feature set and 13-dimensional cepstral feature set.  Table \ref {comprelwercepstral} shows that the conclusions that we drew from these comparisons still hold in the 13-dimensional feature set, namely there is statistical dependence within and between the 13-dimensional frames in the state regions.  Figure \ref {realwithinhistcep} shows a histogram of the per state correlation coefficients between adjacent frames in the state regions.  The analogous histogram for 39-dimensions is displayed in Figure \ref {realwithinhist}.  Both histograms have the same peak at correlation coefficient \(\sim 0.8\), but as expected, the cepstral frames exhibit smaller correlations than the frames with a full feature set.   

\begin{table}
\centering
\begin{tabular}{|l|r|r|}
\hline
code & WER all features & WER cepstral\\
\hline
\hline
\(r_1 r_2 r_3\) & 18 & 30\\
\(r_1 r_1 r_1\) & 24 & 37\\
\(s_1 s_2 s_3\) & 0.6 & 1.4\\
\(s_1 s_1 s_1\) & 4.4 & 13\\
\(r_1 s_2 s_3\) & 5.7 & 9.8\\
\(r_1 s_1 s_1\) & 9.4 & 19 \\
\hline
\hline
\end{tabular}
\caption{Comparison of WER associated with state region operations on the 39-dimensional features and the 13 dimensional features.\label{compabswercepstral}}
\end{table}

\begin{table}
\centering
\begin{tabular}{|c|c|c|}
\hline
 & \multicolumn{2}{|c|}{Relative WER reduction}\\ \cline{2-3}
Comparison & 39 features & 13 features\\
\hline
\hline
\(s_1 s_1 s_1 \to s_1 s_2 s_3\) & 86\% & 89\% \\
\(r_1 r_1 r_1 \to r_1 r_2 r_3\) & 25\% & 19\% \\
\(r_1 r_1 r_1 \to s_1 s_1 s_1\) & 81\% & 65\% \\
\(r_1 s_2 s_3 \to s_1 s_2 s_3\) & 89\% & 86\%\\
\(r_1 s_1 s_1 \to s_1 s_1 s_1\) & 53\% & 31\%\\
\hline
\hline
\end{tabular}
\caption{Comparisons and their relative reduction in WER on the 39-dimensional and the 13-dimensional feature sets.\label{comprelwercepstral}}
\end{table}

\subsection{Variant 2: speaker dependent resampling}

\begin{figure}
\centering
\includegraphics[width=.9\textwidth]{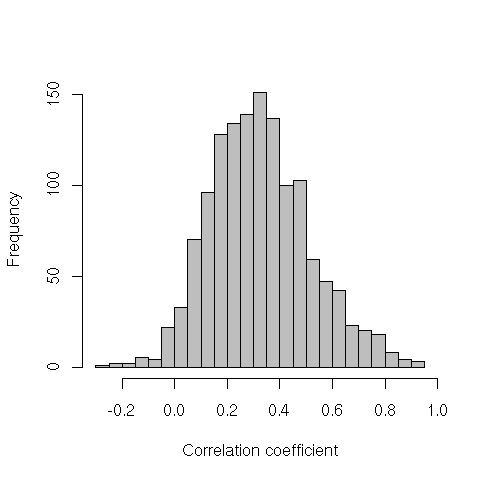}
\caption{Histogram of within state region correlation coefficients on speaker dependent resampled test data.\label {sdresampledwithinhist}}
\end{figure}

Another criticism of the experiments in section~\ref{mainsubsection} is that the pseudo utterances that we created did not respect the underlying speaker structure.  Could our error rates be artificially low because of this?  To try to  answer this we created an alternate version of our test set using speaker dependent resampling of the test data.  The WER on this pseudo test data is 2.0\% with 186 errors as opposed to WER of 0.6\% and 52 errors on the pseudo test data created using speaker independent resampling.  One problem with this experiment is that the real test is small and spread out over 18 speakers, so the urns that we use for speaker dependent resampling have relatively few frames in them.  Could it be that the frames are correlated in this pseudo test data and that this is partly to blame for the higher error rate?  Figure \ref {sdresampledwithinhist} shows a histogram of the within state region correlation coefficients on the speaker dependent resampled test data.  It is clear that there is much more correlation in adjacent frames than one would expect with resampled data (e.g., compare this with Figure \ref {siresampledwithinhist}).  Table \ref {sdspeakercorr} compares the between state region and speaker correlation coefficients measured on speaker independent and speaker dependent resampled test data.  Again the speaker dependent test data shows more between state region correlation.  Clearly, to settle this question conclusively we need to re-examine this question with a corpus with more data per speaker, but in light of the amount of correlation in the data we created, it seems plausible that ignoring the speaker structure when resampling does not create utterances with artificially low error rates.

The alert reader will notice that there is a surprisingly high level of between state region correlation in the pseudo data created by speaker independent resampling and even higher levels in the pseudo data created by speaker dependent resampling.  Shouldn't this be zero by construction?  One possible explanation for why the resampled frames are not fully decorrelated is because of small sample sizes: filling the state urns from the small test set guarantees that many of the state urns contain relatively few frames and, when we draw frames from these urns during the resampling process, this produces sequences of correlated -- even identical -- frames.  In particular, speaker dependent resampling will exacerbate this small sample size effect. Another possible explanation is that we are resampling using real English sentences that consist at the phonetic level of a fairly regular sequence of alternating consonants and vowels with the occasional silence thrown in.  The scores for vowels tend to be lower (better) than the scores for consonants, so this results in a regular sequence of scores: low high low high\ldots  Recall that when we compute the between state region correlation coefficient we select the first score emitted by each state and that each triphone model has three states, so the resulting score sequence will look something like: low low low high high high low low low\ldots Recall also that we compute the correlation coefficient relative to the mean of all of the scores in the speaker's utterances.  This will result in a non-zero correlation.  For example, the correlation coefficient of the adjacent scores in the sequence (-1, -1, -1, 1, 1, 1) is 0.6.  Note that this correlation is an artifact due to our use of the scores as a proxy for the frames.  Just to be certain that this analysis is correct and to exclude the possibility of a bug in our resampling code, we randomized the state sequences that we use to generate the pseudo utterances, then used speaker independent resampling.  The resulting between state region correlation coefficients range between -0.011 and 0.0098.  

\begin{table}
\centering
\begin{tabular}{|c|c|c|c|c|}
\hline
 & \multicolumn{2}{|c|}{Speaker independent} & \multicolumn{2}{|c|}{Speaker dependent}\\ \cline{2-5}
Speaker & \(\rho\) : Between & \(\rho\) : Speaker &\(\rho\) : Between & \(\rho\) : Speaker\\
\hline
\hline
440 & 1.55e-01 & 4.18e-03   & 2.72e-01 & 1.78e-02  \\
441 & 1.90e-01 & 1.86e-03   & 2.67e-01 & 9.50e-02  \\
442 & 1.82e-01 & -1.55e-02  & 2.72e-01 & 1.75e-02  \\
443 & 1.62e-01 & 3.89e-03   & 2.63e-01 & 3.01e-02  \\
444 & 1.62e-01 & 8.73e-03   & 1.93e-01 & 1.25e-03  \\
445 & 1.66e-01 & 1.71e-02   & 2.69e-01 & 9.12e-03  \\
446 & 1.68e-01 & 1.50e-03   & 2.76e-01 & 1.49e-03  \\
447 & 1.52e-01 & -1.44e-02  & 1.97e-01 & 8.73e-03  \\
4k0 & 1.74e-01 & 1.83e-02   & 2.12e-01 & 2.63e-02  \\
4k1 & 1.47e-01 & 2.48e-02   & 2.69e-01 & 4.06e-02  \\
4k2 & 1.59e-01 & -2.29e-03  & 2.43e-01 & 1.61e-02  \\
4k3 & 1.33e-01 & -3.52e-03  & 2.22e-01 & 4.07e-02  \\
4k4 & 1.63e-01 & -3.60e-03  & 2.24e-01 & 1.67e-01  \\
4k6 & 1.52e-01 & 1.65e-02   & 2.24e-01 & -9.88e-03 \\
4k7 & 1.64e-01 & -1.31e-02  & 2.67e-01 & 3.17e-02  \\
4k8 & 1.56e-01 & -8.97e-03  & 2.08e-01 & 2.98e-02  \\
4k9 & 1.38e-01 & 1.68e-02   & 2.21e-01 & 1.17e-02  \\
4ka & 1.15e-01 & 9.14e-03   & 2.10e-01 & 1.73e-01  \\
\hline
\hline
\end{tabular}
\caption{Correlations on speaker dependent resampled data.\label{sdspeakercorr}}
\end{table}

\section{Discussion} \label{discussion}

In this paper we have used novel methods to explore the phenomenology of spoken utterances.  The methods that we employed -- simulation, resampling, and hypothesis testing -- are standard statistical tools but are rarely used in the field of speech recognition.  We hope that this paper highlights their utility and inspires other researchers to use them.  Another novelty in this paper is our use of scores as proxies for features.  Scores have the advantage of being one-dimensional but nevertheless display many interesting properties of the features in relation to the models.    

In section \ref {chisqscores} we show that real data strongly deviates from the diagonal normal output assumption.  This result, while interesting, should not be surprising to any reader.  More surprising, however, is the experiment that shows that when we recognize pseudo test data created by resampling from the test data -- these data obey the HMM generation model but not the diagonal normal output assumption -- the resulting WER is 0.5\%.  This WER is similar to the WER on pseudo test data created by simulating from the model (0.2\%) -- these data satisfy all of the model assumptions.  We deduce that there cannot be that much overlap in either the output distributions or the data, which is also very surprising.   More importantly, however, we conclude that if test data satisfy the Markov assumptions, then they will exhibit low WER whether or not they satisfy the diagonal normal output assumption.  The pseudo data created by resampling are simply a version of the original test data that has been rearranged to satisfy the Markov assumption, but this reordering of the test data results in the WER decreasing from 18\% to 0.5\%.  It follows that real data's departure from the HMM generation model must be the significant source of recognition errors.

In related experiments we demonstrate that real data do exhibit statistical dependence and we do some preliminary investigations into the nature of this dependency.  We find that there is a significant amount of correlation between the acoustic scores in adjacent frames emitted by the same state, but more surprisingly, that there is nearly as much correlation between the leading scores emitted by adjacent states.  On the other hand we can attribute little to none of the statistical dependency to speaker effects.  This may be in part due to our use of the WSJ corpus.  We suspect that more spontaneous speech corpora, e.g. Fisher, may exhibit a richer dependency structure. 

It is important to note we are careful to say ``if test data satisfy the Markov assumptions, then they will exhibit low WER whether or not they satisfy the diagonal normal output assumption''.  In particular, we are not claiming that using better acoustic models will not result in fewer errors.  A better acoustic model will result in more accurate state classification which in turn will result in better recognition performance.  Indeed, if we compare our baseline unimodal models with 10 component mixture models, then the WER goes from 18\% to 12\%.  Using larger and more sophisticated models (crossword triphones, skipping, and a transformation to ameliorate the diagonal covariance assumption) bring the WER to under 6\%.  However, we do believe that this error rate is close to the lowest WER that one could achieve on this test set using a single HMM-based model (i.e. not combining different recognizer outputs \emph{a la} ROVER) trained using maximum likelihood.   We also believe that the residual error rate of 5\% is due to the statistical dependency in the data that is not accounted for in the model.

We are not the first researchers to show that frames are correlated, nor are we the first to suggest that one could model speech better by using models other than HMMs.  However, previous work has has concentrated on using intuition gleaned from studies of data to create more elaborate models, with their correspondingly more elaborate estimation procedures, and then using WER to judge the value of the new model. In our view a critical step has been left out of this process, namely, a more fundamental (than WER) analysis of how well the data fit the model.  We, on the other hand, start with the questions: (a) what properties do real data exhibit that violate our current model assumptions, (b) what effect does these properties have on recognition performance, and (c) are there aspects of these properties that we can reliably model?  In this paper we have begun to answer (a) and (b).  In future work we shall study (c). For example, when a recognizer computes the score for a hypothesis it treats the scores emitted by the underlying state sequence as independent and just adds them up.  However we have shown that scores are in fact highly correlated and we suspect that the amount of correlation varies widely across states, speakers, and speaking modes.  If this is true and we could reliably model the joint distribution of scores, then it ought to lead to significant improvements in recognition performance.  In fact, we believe that MMI is compensating for exactly this effect, but in a very crude and indirect way.


\begin{thebibliography}{99}
\bibitem{abramowitz} Abramowitz, M., and Stegun, I. A., eds. (1972). \emph{Handbook of Mathematical Functions with Formulas, Graphs, and Mathematical Tables}, Dover, New York.
\bibitem{baker} Baker, J. K. (1975). The DRAGON system -- an overview. \emph{IEEE Transactions on Acoustics, Speech, and Signal Processing}, \textbf{ASSP-23} (1), 24-29.
\bibitem{bilmes} Bilmes, J. A. (2003). Buried Markov models: a graphical-modeling approach to automatic speech recognition. \emph{Computer Speech and Language}, \textbf{17} (2-3), 213-231.
\bibitem{chen} Chen, S., Kingsbury, B., Mangu, L., Povey, D., Saon, G., Soltau, H., and Zweig, G. (2006). Advances in speech transcription at IBM under the DARPA EARS program. \emph{IEEE Transactions on Audio, Speech, and Language Processing}, \textbf{14} (5), 1596-1608. 
\bibitem{efron1} Efron, B. (1979). Bootstrap methods: another look at the jackknife. \emph{The Annals of Statistics}, \textbf{7} (1), 1-26.
\bibitem{efron2} Efron, B. (1982). \emph{The Jackknife, the Bootstrap and Other Resampling Plans}, SIAM CBMS-NSF Monographs, 38.
\bibitem{jelinek} Jelinek, F. (1976). Continuous speech recognition by statistical methods. \emph{IEEE Proceedings}, \textbf{64} (4), 532-556.
\bibitem{kflee} Lee, K.-F. (1988). \emph{Large-Vocabulary Speaker-Independent Continuous Speech Recognition: the Sphinx System}, Ph. D. Thesis, Carnegie Mellon University.
\bibitem{lippmann} Lippmann, R. (1997). Speech recognition by machines and humans. \emph{Speech Communication}, \textbf{22}, 1-15.
\bibitem{mcallaster} McAllaster, D., Gillick, L., Scattone, F., and Newman, M.  (1998). Studies with fabricated Switchboard data: exploring sources of model-data mismatch. \emph{Proceedings of DARPA Broadcast News Transcription and Understanding Workshop}, 306-310.
\bibitem{morgan} Morgan, N., Zhu, Q., Stolcke, A., Sonmez, K., Sivadas, S., Shinozaki, T., Ostendorf, M., Jain, P., Hermansky, H., Ellis, D., Doddington, G., Chen, B., Cretin, O., Bourlard, H., and Athineos, M. (2005). Pushing the envelope - aside : beyond the spectral envelope as the fundamental representation for speech recognition.  \emph{IEEE Signal Processing Magazine}, \textbf{22} (5), 81-88.
\bibitem{ostendorf} Ostendorf, M., Digalakis, V., and Kimball, O. A. (1996). From HMM's to segment models: a unified view of stochastic modeling for speech recognition. \emph{IEEE Transactions on Speech and Audio Processing}, \textbf{4} (5), 360-378.
\bibitem{povey2} Povey, D. (2004). \emph{Discriminative Training for Large Vocabulary Speech Recognition}, Ph. D. Thesis, Cambridge University.
\bibitem{woodland1} Woodland, P. C., Odell, J. J., Valtchev, V., and Young, S. J. (1994).  Large vocabulary continuous speech recognition using HTK.  \emph{Proceedings of ICASSP-94}, 306-310.
\bibitem{woodland2} Woodland, P. C., and Povey, D. (2002). Large scale discriminative training of hidden Markov models for speech recognition. \emph{Computer Speech and Language}, \textbf{16} (1), 25-47.
\bibitem{young} Young, S. J., Evermann, G., Gales, M. J. F., Kershaw, D., Moore, G., Odell, J. J., Ollason, D. G., Povey, D., Valtchev, V., and Woodland,. P. C. (2006). \emph{The HTK Book Version 3.4}, Manual, Cambridge University Engineering Department.
\bibitem{zue} Zue, V., Glass, J., Philips, M., and Seneff, S, (1989). Acoustic segmentation and phonetic classification in the SUMMIT system. \emph{Proceedings of ICASSP-89}, 389-392.
\end{thebibliography}
\end{document}